\documentclass{INTERSPEECH2023}

\usepackage{stfloats}
\usepackage{caption}
\usepackage{amsfonts}
\usepackage{multirow}

\interspeechcameraready

\title{Towards Rehearsal-Free Multilingual ASR: A LoRA-based Case Study on Whisper }

\name{Tianyi Xu$^1$, Kaixun Huang$^1$, Pengcheng Guo$^1$, Yu Zhou$^2$, Longtao Huang$^2$, Hui Xue$^2$, Lei Xie$^\dagger$}
\address{
  $^1$ Audio, Speech and Language Processing Group (ASLP@NPU), School of Computer Science, \\ Northwestern Polytechnical University, Xi'an, China
  \\ $^2$ Alibaba Group, China
\email{xutianyi@mail.nwpu.edu.cn, lxie@nwpu.edu.cn}
\thanks{ $^\dagger$Lei Xie is the corresponding author.}}
\usepackage{threeparttable}
\begin{document}
\maketitle
\begin{abstract}
  Pre-trained multilingual speech foundation models, like Whisper, have shown impressive performance across different languages. However, adapting these models to new or specific languages is computationally extensive and faces catastrophic forgetting problems. Addressing these issues, our study investigates strategies to enhance the model on new languages in the absence of original training data, while also preserving the established performance on the original languages.  Specifically, we first compare various LoRA-based methods to find out their vulnerability to forgetting. To mitigate this issue, we propose to leverage the LoRA parameters from the original model for approximate orthogonal gradient descent on the new samples. Additionally, we also introduce a learnable rank coefficient to allocate trainable parameters for more efficient training. Our experiments with a Chinese Whisper model (for Uyghur and Tibetan) yield better results with a more compact parameter set. 

\end{abstract}

\section{Introduction}

\begin{figure*}[t]
\centering

\resizebox{0.9\textwidth}{!}{\includegraphics{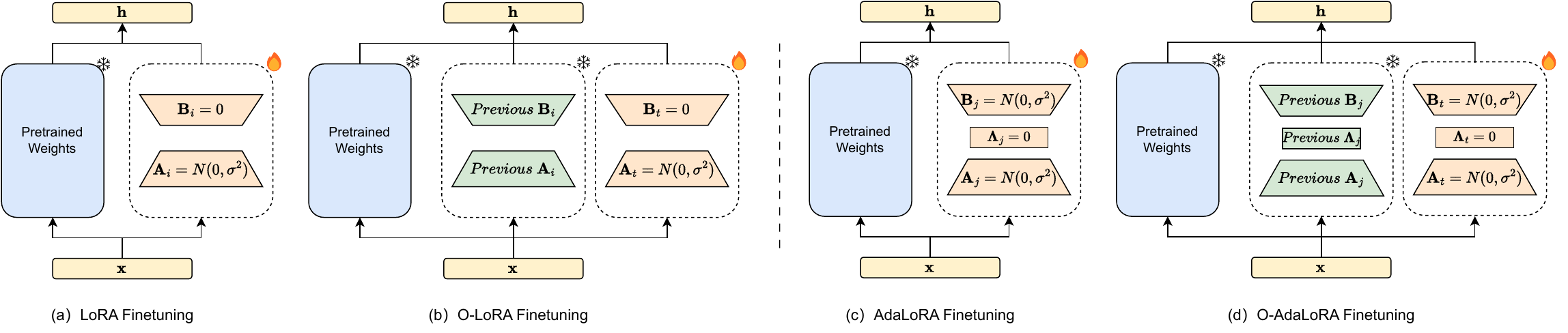}}
\caption{The structure of O-LoRA and O-AdaLoRA.}
\vspace{-0.5cm}
\label{lora}
\end{figure*}
Recently, a large number of pre-trained foundation models have emerged in the speech areas, either trained in a self-supervised~\cite{hsu2021hubert,zhang2023google} or supervised learning~\cite{radford2023robust,chu2023qwen} manner. Among these models, OpenAI's Whisper~\cite{radford2023robust} has shown promising results for multilingual automatic speech recognition (ASR) and translation tasks. Despite its robust performance across various languages, Whisper exhibits suboptimal recognition capabilities for low-resource languages, including multilingual Tibetan and Uyghur. The common approach to improve performance in such languages involves fine-tuning the pre-trained model with data from the target language, either with full fine-tuing, or with Lora~\cite{hu2022lora}, a popular parameter efficient tuning method. Both approach, however, comes with a significant drawback. The fine-tuned model often experiences a noticeable decline in its ability to recognize the languages it was originally trained in, leading to a phenomenon known as catastrophic forgetting. 

To counteract catastrophic forgetting, many continual learning methods have been proposed in the image classification community. Existing studies about continual learning can be mainly classified into rehearsal-based, regularization-based, and architecture-based approaches. Rehearsal-based approaches~\cite{chaudhry2019tiny,rolnick2019experience} involve storing samples from previous tasks and combining them with the new data to jointly train the model. However, this method may not always be practical when original samples are unavailable or privacy concerns arise. For large-scale foundation models, the significant memory needed to store these samples also poses a challenge. Regularization-based approaches~\cite{zenke2017continual,aljundi2018memory,kirkpatrick2017overcoming} add a regularization term to the loss function to discourage large changes to the weights trained on previous tasks. While this method helps maintain training stability, it may limit the model's flexibility to learn new tasks. Architecture-based approaches add task-specific adapters for the new tasks, which prevents modifications to the original parameters~\cite{wang2023rehearsal,razdaibiedina2023progressive}. However, this requires the model to infer with the new task-id to decide which part of the model should be used, limiting its applicability to foundation models that lack access to task-ids.

In ASR, research on continual learning is relatively scarce~\cite{vander2023using,yang2022online}. Existing studies typically perform direct gradient updates on the model's hidden layers within a common vector space for all tasks. However, recent research in the field of Natural Language Processing (NLP) has proposed a promising alternative. When training on new tasks, the loss gradients for new samples are projected orthogonally to the gradient subspace of past samples~\cite{farajtabar2020orthogonal}. Given that neural networks are often over-parameterized, this strategy avoids conflicts with previous loss functions, thereby mitigating the problem of catastrophic forgetting. O-LoRA~\cite{wang2023orthogonal} has indicated that the LoRA parameters not only capture the gradient subspace of previous tasks, but can also be used to approximate the gradients of previous tasks to guide continual learning. 

Our main contribution is investigating the catastrophic forgetting phenomenon observed in Whisper fine-tunig and utilizing orthogonal gradient descent to alleviate it. Our experimental result shows that the LoRA parameters of the original models are good representatives of the original data samples, and can be used to instruct the continual learning possess. Furthermore, inspired by the AdaLoRA~\cite{zhang2023adaptive}, we utilize a dynamic approach to allocate the ranks of matrices more judiciously, thereby improving model performance and further reducing the parameter count necessary for training.

Distinct from existing papers on continual learning research in ASR, this study emphasizes addressing a practical challenge faced by pre-trained speech foundation models: \textit{How to fine-tune large pre-trained models for new tasks in a parameter-efficient way when the original training data is not available.} In real-world applications, new data would accumulate over time and these data could potentially be used to improve the model performance. Furthermore, new requirements would emerge and demand the model to solve new tasks. Our proposed method offers the following advantages: (1) \textbf{Rehearsal-free}:  In many scenarios, the new data would be much less than the data used to train the original model, therefore it is not efficient to retrain the model using all data. Furthermore, due to privacy concerns, it may not be possible to retain old data. (2) \textbf{ Parameter-efficient}: Given that the data for new tasks may be limited and insufficient for full-finetuning, the proposed method should be parameter-efficient. We use the LoRA method to perform parameter-efficient fintuning and borrow ideas from AdaLoRA to dynamically assign a rank to LoRA parameters to further reduce the additional parameters. (3) \textbf{Task-id-free}: For real-world applications, such as multilingual ASR, we aim for the model to infer without explicitly task id, enhancing its generalizability.


To the best of our knowledge, our work is the first to investigate a rehearsal-free method based on orthogonal-gradient decent to solve continual learning challenges encountered by ASR models like Whisper.
Although we primarily discuss continual learning for the Whisper model in this study, the proposed methods can be expanded to other speech foundation models with the Transformer architecture. 


\section{Method }


\subsection{Problem definition}
First, we provide a more formal definition of the domain transfer problem in multilingual ASR models. We begin by assuming an initial model, $M_0$, which was trained with LoRA on a dataset $D_0$. Subsequent continual learning is built upon $M_0$, with an intention to preserve its capabilities, while $D_0$ is no longer usable. Afterward, we have a series of labeled datasets of various languages, $D=\{{D_n}\}^N_{n=1}$, intended for subsequent multilingual fine-tuning training. The model trained with $D_n$ is referred to as $M_n$. $N$ represents the total number of subsets, and the size of each $D_n$ is significantly smaller than $D_0$. Utilizing $D_0$ for review training or retraining with it requires a substantial training cost. For an ideal continual learning model, at each time step $n$, we desire it to exhibit the following characteristics:

\begin{enumerate}
    \item \textbf{No performance degradation on previous tasks}: $M_n$ achieves the same or better results on $D_{n-1}$ compared to $M_{n-1}$.
    \item \textbf{Improved performance on new tasks}: $M_n$ achieves better results on $D_n$ compared to $M_{n-1}$.
    \item \textbf{Comparable performance to a model trained on all data}: $M_n$ achieves comparable results to a model trained from scratch using the datasets $D=\{{D_n}\}^N_{n=1}$.
    \item \textbf{Parameter efficiency}: The required training expenditure is significantly less than retraining the model using all datasets $D=\{{D_n}\}^N_{n=1}$.
\end{enumerate}
 
\subsection{
LoRA}

To achieve parameter efficiency, we use LoRA~\cite{hu2022lora}, which was first proposed in NLP, to adapt models to specific domains or downstream tasks. Research has indicated that the weights of pre-trained large language models often occupy a low intrinsic dimensional space. Inspired by this observation, LoRA freezes the original weights and updates only the low-rank incremental weight matrices. Specifically, consider the $i$-th forward $f_i(\textbf{x})=\textbf{x}\textbf{W}^\intercal+\textbf{b}_i$ where $\textbf{W}_i\in\mathbb{R}^{d_1 \times d_2}$, and $\textbf{b}_i\in\mathbb{R}^{d_2}$ denotes the frozen weight and bias, as shown in figure \ref{lora} (a). LoRA constrains its updates by representing them as a low-rank decomposition, specifically, the forward is mwddified as:
\begin{equation}
f_i(\textbf{x})=\textbf{x}(\textbf{W}_{i}+\Delta \textbf{W}_i)^\intercal+\textbf{b}_i \Delta \textbf{W}_i=\textbf{A}_i\textbf{B}_i
\end{equation}
 where $\textbf{A}_i\in\mathbb{R}^{{d_1}\times r}$ $\textbf{B}_i\in\mathbb{R}^{{r}\times d_2}$, and the rank $r\ll min\{d_1,d_2\}$, $\textbf{W}_i$ remains fixed during training and does not receive gradient updates, whereas $\textbf{A}_i$ and $\textbf{B}_i$
contain trainable parameters. 
 
\subsection{Continual learning with O-LoRA}
We base our method on O-LoRA~\cite{wang2023orthogonal}. O-LoRA incrementally learns new tasks in a direction orthogonal to the LoRA subspace of past tasks while fixing the previous parameters. For each task, we introduce a set of new LoRA parameters dimensionally identical to the old ones denoted as { ${\textbf{A}}_t$, ${\textbf{B}}_t$}, where  $\textbf{A}_t\in\mathbb{R}^{{d_1}\times r}$ $\textbf{B}_t\in\mathbb{R}^{{r}\times d_2}$ and $r\ll min\{d_1,d_2\}$. We approximate the parameter update subspace ${\mathcal{U}}_t$ for the $t$-th task as the subspace spanned by the column vectors of ${\textbf{A}}_t$, as shown in figure \ref{lora} (b).  Mathematically, to ensure the orthogonality between the subspace  ${\mathcal{U}}$ and the subspace  ${\mathcal{W}}$, we need to satisfy:
\begin{equation}
<u,w>=0, \forall u \in \mathcal{U}, \forall w \in \mathcal{W}
\end{equation}
Therefore, achieving orthogonality between the LoRA subspaces of task $i$   (${\mathcal{U}}_i$) and task $t$ (${\mathcal{U}}_t$) can be expressed as:
\begin{equation}
O_{i,t}=\textbf{A}^\intercal_i\textbf{A}_t=0
\end{equation}
We add an orthogonal loss to enforce this orthogonality:
\begin{equation}
L_{orth}(A_i,A_t)=\sum\limits_{j,k} ||O_{i,t}[j,k]||^2
\end{equation}
where $O_{i,t}[j,k]$denotes the element at the $j$-th row and $k$-th column of  $O_{i,t}$ .
Finally our training objective is defined as:
\begin{equation}
L=L_{whisper}+\lambda_1 L_{orth}(A_i,A_t)
\end{equation}
where $\lambda_1 $ is the weight of the orthogonality loss. During the training process, we fix the previous LoRA parameters \{$A_i,B_i|i<t$\} to mitigate forgetting of past knowledge. 

\subsection{Orthognal adaptive low-rank adaptation (O-AdaLoRA)}

The importance of weight parameters to the performance varies across different layers and modules. Intuitively, some weight matrices should be prioritized higher in the tuning process. However, the traditional LoRA approach takes a one-size-fits-all strategy by assigning a uniform rank to all weight matrices, which does not account for the possibility that the relative importance of weights may change depending on the context, potentially leading to suboptimal performance. To address this, the AdaLoRA~\cite{zhang2023adaptive} (Adaptive Low-Rank Adaptation) introduces a dynamic rank allocation method that is sensitive to the importance of weights. AdaLoRA makes two key improvements to the original LoRA mechanism: Firstly, it employs a parameterization strategy based on Singular Value Decomposition (SVD) to represent the incremental update term $\Delta \textbf{W}_j$ as:
\begin{equation}
\Delta \textbf{W}_j={\textbf{A}_j\boldsymbol\Lambda_j} {\textbf{B}_j} ; {\textbf{A}_j}^\intercal{\textbf{A}_j}={\textbf{B}_j}{\textbf{B}_j}^\intercal=\textbf{I}
\end{equation}
where ${ \textbf{A}_j}\in\mathbb{R}^{d_1  \times r}$ and ${ \textbf{B}_j}\in\mathbb{R}^{r  \times d_2}$contain the left/right singular vectors of :$\Delta \textbf{W}_j$ . $\textbf{I}$ denotes the identity matrix. and the {${\textbf{A}}_j$,${\textbf{B}}_j$} are enforced to be orthogonal matrices with a regularization term: 
\begin{equation}
L_{adalora}(A_j,B_j)=|| {\textbf{A}_j}^\intercal{\textbf{A}_j}-\textbf{I}||^2+|| {\textbf{B}_j}{\textbf{B}_j}^\intercal-\textbf{I}||^2
\end{equation}

The diagonal matrix $\boldsymbol{\Lambda}_j\in\mathbb{R}^{r  \times r}$ contains $r$ singular values, with  $r\ll min\{d_1,d_2\}$. Only ${\textbf{A}}_j$, $\boldsymbol\Lambda_j$,  ${\textbf{B}}_j$ are trainable, as shown in figure \ref{lora} (c). 
Second, in adaptation, AdaLoRA dynamically allocates an overall rank budget to its update matrices {$\Delta \textbf{W}_j$}. This is achieved by iteratively masking out less important singular values after every gradient update step. A sensitivity-based importance metric is utilized to measure and sort the importance of the $k$-th triplet { $\boldsymbol{\Lambda}_j^{k,k}$, ${\textbf{B}}_j^{*,k}$, ${\textbf{A}}_j^{k,*}$} of the $j$-th weight matrix $W_j$, which takes account of both singular values and vectors. The non-zero  $\boldsymbol{\Lambda}_j^{k,k}$acts as the rank coefficient to control the allocated rank budget.

To learn new tasks in a direction orthogonal to the past tasks, we just need to ensure the orthogonality between the subspace of $j$-th task $\mathcal{U}_j$ and the subspace of $t$-th task  $\mathcal{U}_t$. Like section 2.3, we add a new set of $\{\textbf{A}_t$,$\textbf{B}_t$,$\boldsymbol{\Lambda}_t\}$ as shown in figure \ref{lora} (d),  and enforce the orthogonality with:
\begin{equation}
O_{j,t}=\textbf{A}^\intercal_j\textbf{A}_t=0
\end{equation}
Similar to equation (5), our training objective is defined as:
\begin{equation}
L=L_{whisper}+\lambda_1 L_{orth}(\textbf{A}_j,\textbf{A}_t)+\lambda_2 L_{adalora}(\textbf{A}_t,\textbf{B}_t)
\end{equation}
where $\lambda_1$ and $\lambda_2$ are regularization coefficients. Since combining all $\{\textbf{A}_j$,$\textbf{B}_j$,$\boldsymbol{\Lambda}_j|j<t\}$ before the forward operation is mathematically equivalent to computing each one individually, the additional computational overhead is minimal.
The additional memory usage can be mitigated by merging the updates corresponding to the LoRA parameters into the initial ones.
\begin{table}[!t]

\centering
\caption{Datasets used in this study.}
\label{dataset}
\resizebox{0.47\textwidth}{!}{
\begin{tabular}{@{}lll@{}}
\toprule
Language & Source Dataset                                             & Hours of training audio \\ \midrule
Chinese  & Wenetspeech,Aishell-1                                        & 10000                   \\
Uyghur   & CommonVoice,Thuyg20, In-house Uyghur Dataset & 250                     \\
Tibetan  & XBMU-AMDO31,TIBMD, In-house Tibetan Dataset           & 215                     \\ \bottomrule
\end{tabular}
}
\vspace{-0.5cm} 
\end{table}

\begin{table*}[h]
\centering
\setlength{\belowcaptionskip}{0.05cm} 
\setlength{\abovecaptionskip}{0.15cm} 
\caption{Peformance (WER/CER\%)  after Uyghur finetuning, N/A indicates the model can not transcribe in the correct language. A WER larger than 100 is still in the correct language.}
\label{uyghur}
\resizebox{0.85\textwidth}{!}{
\begin{tabular}{l|ll|ll|ll|ll|ll}
\hline
                  & \multicolumn{2}{l|}{orignal-v2} & \multicolumn{2}{l|}{full-ft} & \multicolumn{2}{l|}{LoRA} & \multicolumn{2}{l|}{AdaLoRA} & \multicolumn{2}{l}{AdaLoRA +Freeze encoder} \\
                  & base            & large         & base          & large        & base        & large       & base          & large        & base                 & large                \\ \hline
Aishell-1        & 22.20           & 8.82          & N/A& 30.87        & N/A& 24.31       & N/A& 28.45        & N/A& 25.40                \\
In-house Manderin & 89.32           & 47.94         & N/A& N/A& N/A& N/A& N/A& N/A& N/A& N/A\\
Thuyg-20          & 147.72          & 99.92         & 51.74         & 41.17        & 44.68       & 39.45       & 37.76         & 33.63        & 38.32                & 35.89                \\ \hline
\end{tabular}
}
\vspace{-0.5cm} 
\end{table*}

\begin{table}[!t]
\centering
\caption{Performance (CER\%) of Whisper large-v2 finetuned on Chinese dataset.}
\label{chinese}
\resizebox{0.47\textwidth}{!}{
\begin{tabular}{@{}lllll@{}}
\toprule
                   & original large-v2 & full-ft & LoRA   & AdaLoRA \\ \midrule
Aishell-1            & 8.81& 5.72    & 4.23   & 3.78    \\
In house Chinese& 47.94             & 40.57   & 36.98  & 34.43   \\
\#Trainable Params & -                 & 100\%   & 3.47\% & 1.36\%  \\ \bottomrule
\end{tabular}
}
\vspace{-0.3cm} 
\end{table}

\section{Experiments}

\subsection{Data}

In our study, we trained speech recognition models for Chinese, Uyghur, and Tibetian using a variety of datasets. For Chinese, we use the WenetSpeech~\cite{zhang2022wenetspeech} and Aishell-1~\cite{bu2017Aishell} datasets for training. For testing, we use the Aishell-1 test set and an in-house Chinese test set from a voice assistant product with more complex acoustic environments. For Uyghur, we use the Common Voice Dataset 15.0~\cite{ardila2019common} and THUYG-20~\cite{ THUGY20_2015}, using the test set from THUYG-20 for evaluation. As for Tibetan, we use the TIBMD~\cite{zhao2020open} and XBMU-AMDO31 datasets~\cite{XBMU-AMDO31}, and we use TIBMD's test set for testing. We did not differentiate between the dialects in the TIBMD dataset. For mixed dataset training, the mixture is weighted according to the number of elements in each component while we down-weighted larger datasets. Due to the limited availability of open-source datasets for Tibetan and Uyghur, we add approximately 100 hours of in-house data for each language.  A detailed list of our data sources for each language can be found in the referenced table \ref{dataset}.

 We noticed that when processing Uyghur and Tibetan, Whisper output words in a Romanized form—written in the Latin alphabet, instead of their original characters. We hypothesize that Whisper's training included Romanized versions of these languages. To keep our work consistent with Whisper's training, we use the uroman tool~\cite{hermjakob-etal-2018-box} to Romanize our transcripts before training and testing. Also, we stripped out all punctuation from these transcripts, and we discounted any punctuation that appeared in the output when we measured our results.
Since Whisper doesn't include a specific language ID token for Uyghur, we use the language label for Uzbek, which is linguistically similar.

\subsection{Experimental setups}
All our experiments were performed using the Parameter-Efficient Fine-Tuning (PEFT) library in Hugging Face. To assess the performance with different parameter sizes, we investigated two sizes of the Whisper model: the base model with 74 M parameters and the large model with 1,550 M parameters, we use the large-v2 model unless specified otherwise. For PEFT, we apply low-rank updates to a set of matrices: {$W_q$, $W_k$, $W_v$, $W_o$, $W_{fc1}$, $W_{fc2}$}. Following the parameter config of ~\cite{liu2023sparsely}, for LoRA and O-LoRA, we use a rank $r=32$ for matrix updates. For AdaLoRA and O-AdaLoRA, we set the initial rank $r=12$ and a target rank of 8. Inspired by ~\cite{winata2023overcoming}, we set the learning rate for the PEFT model at 1e-3 for the initial task and reduced it to 1e-4 for subsequent tasks. For full-model finetuning, we use a learning rate of 1e-5. Whisper's architecture pads audio inputs to 30 seconds, which significantly increases memory usage, leading us to adopt a $batch\_size=2$, with a gradient accumulation of 128 steps. The orthogonal loss coefficient $\lambda_{1,2}=0.5$ is used. For evaluation metrics, we measure performance in Chinese using the Character Error Rate (CER) and for the Romanized versions of Tibetan and Uyghur, we use the Word Error Rate (WER).

\subsection{Comparison with other methods}
We evaluate O-LoRA and O-AdaLoRA against 5 baseline methods. Since we are investigating methods that are rehearsal-free and task-id-free, rehearsal-based methods like A-GEM~\cite{yang2022online}, and adapter-based methods~\cite{vander2023using} that require task-ids were not included. Since we found AdaloRA fine-tuning performs the best, we use AdaLoRA for Multilingual fine-tuning and Monolingual finetuning as toplines. We compared with the following methods: (1)\textbf{SeqFT} (Sequential Finetuning): Full fine-tuning on all tasks in sequence. (2) \textbf{SeqLoRA} (Sequential LoRA Finetuning): Sequentially training a fixed LoRA matrix with data from different languages, the LoRA matrix is merged after every language. (3) \textbf{LwF}(Learning without Forgetting): Regularization-based method that constrains the shared representation layer to be similar to its original state before learning the new task. LWF was the best memory-free and task-id-free continual learning method from~\cite{chang2021towards,vander2022continual,vander2023using}. (4) \textbf{Multi} (Multilingual fintuning): Training with a mixed dataset with all three language datasets with AdaLoRA simultaneously. (5) \textbf{Mono} (Monoilingual fintuning): Fintuning separate whisper models with AdaLoRA using datasets from individual languages. 


\subsection{Limitation of vanilla fine-tuning}

\begin{figure}[t]
  \centering
  \includegraphics[width=0.85\linewidth]{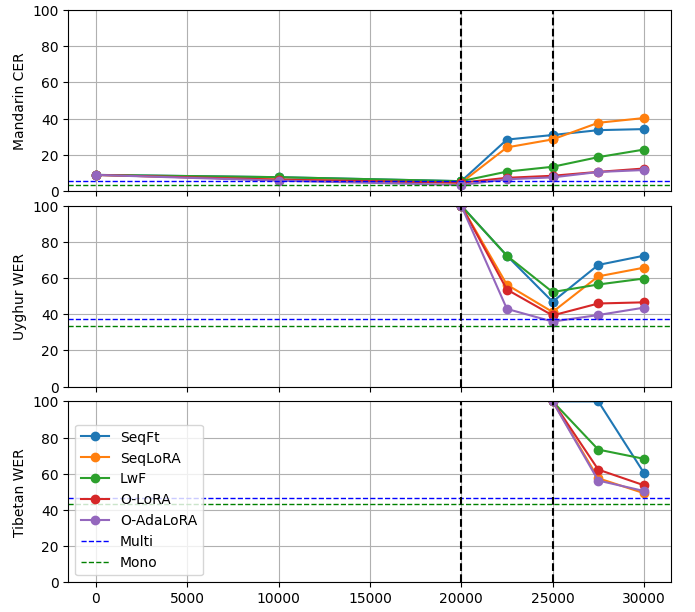}
  \caption{Performance (WER/CER\%) of different methods after transfer between languages}
  \label{orth}
\vspace{-0.5cm}
\end{figure}

We first investigate the possibility of adding a new language (Uyghur) to Whisper across various sizes with baseline approaches: full finetuning, LoRA, AdaLoRA, and freeze encoder approach seen in LLM adaptations. We finetuned Whisper models of different capacities with a dataset composed of Uyghur speech. For this experiment, our language of concern is Chinese and Uyghur. We measure the performance gain in Uyghur with Thuyg-20 while measuring the performance degradation in Chinese with Aishell-1. The findings are detailed in table \ref{uyghur}.
The smaller Whisper models showed a lower tolerance to catastrophic forgetting, as the base model recognized the  Chinese Aishell-1 dataset as Uyghur. A more challenging dataset is found to be more sensitive to catastrophic forgetting, as the Whisper model suffering from catastrophic forgetting can still transcribe Aishell-1 into Chinese, yet the in-house Chinese dataset is being transcribed as Uyghur. While the freeze encoder strategy has the benefit of achieving a similar WER with fewer trainable parameters, it did not have a significant impact on the catastrophic forgetting phenomenon experienced when adding a new language. 

\subsection{Effectiveness of orthogonal subspace learning}

Next, we assessed the effectiveness of orthogonal methods. Unfortunately, since the original Whisper was not trained using LoRA, it does not retain LoRA matrices, and therefore we cannot directly apply our method to protect all languages in Whisper during subsequent fine-tuning. As a compromise, we procure a Chinese model trained using LoRA. We trained a model using the Chinese dataset discussed in Section 3.1 as our initial model $M_0$. The result is shown in table \ref{chinese} and figure \ref{orth}.
Despite having the most trainable parameters, we found that the full-model fine-tuning approach yielded the poorest performance. We speculate this may be due to that our dataset was significantly smaller than the 680,000 hours of data used to train Whisper thus the model struggled to converge. The LoRA model showed slightly inferior results compared to AdaLoRA. We hypothesize that this could be related to the findings in ~\cite{liu2023sparsely}, where it's suggested that $W_v$,$W_o$ should have more trainable parameters than $W_q$,$W_k$ during the finetuning.

We conducted further research into the phenomena of catastrophic forgetting using this Chinese-finetuned Whisper-large model with sequential language training. After initially training on Chinese data for 20,000 steps, we subsequently incorporated Uyghur and Tibetan datasets for an additional 5,000 steps each, using various methods.  Figure \ref{orth} shows the changes in WER/CER.

 Although Whisper's paper~\cite{radford2023robust} suggested that mixed language training may lead to improvements over single-language performance, we found that the three monolingual models performed better than the multi-language model. We suspect this discrepancy arises because our dataset was not as large as Whisper's and the LoRA matrix did not learn the ability to integrate knowledge across languages.
The orthogonal gradient-based approach O-LoRA significantly reduced the catastrophic forgetting issue, with the SeqFT and SeqLoRA having a final Chinese WER after Tibetian fine-tuning of over 30\%. And outperformed the SOTA method LwF that did not require review or task-id, with a lower final Chinese WER (22.82\% vs 12.35\%). Our O-AdaLoRA sped up model convergence and achieved lower WER (53.62\% vs 50.45\%) in the Tibetian fine-tuned result, with a smaller parameter size (3.47\% vs 1.36\%) while having a similar performance retaining the previous tasks.


\section{Conclusions}

This paper takes the first step in investigating fine-tuning Whisper from a continual learning point-of-view and using orthogonal gradient descent to adapt the pre-trained Whisper model for Uyghur and Tibetan languages. We investigated various fine-tuning methods and examined their susceptibility to catastrophic forgetting. We have demonstrated that the LoRA matrices of a model can be used to guide subsequent continual learning in ASR. We proposed to allocate the parameters with a learnable rank coefficient to improve the overall performance. This enables the fine-tuning of foundation models that are entirely deprived of pre-training data while avoiding previously acquired language knowledge being overwritten, greatly enhancing the potential reusability of pre-trained models.

\nocite{*}
\bibliography{mybib.bib} 
\bibliographystyle{IEEEtran} 
\end{document}